\documentclass[10pt,twocolumn,letterpaper]{article}

\usepackage{cvpr}              

\newcommand{\shortcite}[1]{(\citeyear{#1})}

\usepackage{pifont} 

\usepackage[table]{xcolor}   
\usepackage{booktabs}        
\definecolor{bestred}{RGB}{255,220,220}
\definecolor{secondorange}{RGB}{255,236,214}
\definecolor{thirdyellow}{RGB}{255,247,200}

\newcommand{\best}[1]{\cellcolor{bestred}\textbf{#1}}
\newcommand{\second}[1]{\cellcolor{secondorange}{#1}}

\definecolor{cvprblue}{rgb}{0.21,0.49,0.74}
\usepackage[pagebackref,breaklinks,colorlinks,allcolors=cvprblue]{hyperref}

\def\paperID{} 
\def\confName{CVPR}
\def\confYear{2026}

\title{MVGSR: Multi-View Consistent 3D Gaussian Super-Resolution via Epipolar Guidance}

\author{Kaizhe Zhang\\
Xi'an Jiaotong University\\
{\tt\small zkz1081@stu.xjtu.edu.cn}
\and
Shinan Chen\\
Xi'an Jiaotong University\\
{\tt\small chensn@stu.xjtu.edu.cn}
\and
Qian Zhao\\
Xi'an Jiaotong University\\
\and
Weizhan Zhang\\
Xi'an Jiaotong University\\
\and
Caixia Yan\\
Xi'an Jiaotong University\\
\and
Yudeng Xin\\
University of Melbourne\\
}

\begin{document}
\maketitle
\begin{abstract}
Scenes reconstructed by 3D Gaussian Splatting (3DGS) trained on low-resolution (LR) images are unsuitable for high-resolution (HR) rendering. Consequently, a 3DGS super-resolution (SR) method is needed to bridge LR inputs and HR rendering. Early 3DGS SR methods rely on single-image SR networks, which lack cross-view consistency and fail to fuse complementary information across views. More recent video-based SR approaches attempt to address this limitation but require strictly sequential frames, limiting their applicability to unstructured multi-view datasets. In this work, we introduce Multi-View Consistent 3D Gaussian Splatting Super-Resolution (MVGSR), a framework that focuses on integrating multi-view information for 3DGS rendering with high-frequency details and enhanced consistency. We first propose an Auxiliary View Selection Method based on camera poses, making our method adaptable for arbitrarily organized multi-view datasets without the need of temporal continuity or data reordering. Furthermore, we introduce, for the first time, an epipolar-constrained multi-view attention mechanism into 3DGS SR, which serves as the core of our proposed multi-view SR network. This design enables the model to selectively aggregate consistent information from auxiliary views, enhancing the geometric consistency and detail fidelity of 3DGS representations. Extensive experiments demonstrate that our method achieves state-of-the-art performance on both object-centric and scene-level 3DGS SR benchmarks.
\end{abstract}    
\section{Introduction}
\label{sec:intro}

Novel view synthesis has been regarded as a fundamental problem in 3D reconstruction, image editing, and virtual scene navigation, aiming to generate images from unseen viewpoints based on existing observations. Recently, Neural Radiance Fields (NeRF)~\cite{mildenhall2021nerf} and 3D Gaussian Splatting (3DGS)~\cite{kerbl20233d} have achieved remarkable progress in this area. NeRF uses implicit neural representations to generate photorealistic images but struggles with scalability and efficiency, despite recent advances~\cite{chen2022tensorf, fridovich2022plenoxels, garbin2021fastnerf, niemeyer2022regnerf}. In contrast, 3DGS models scenes with Gaussian primitives, enabling a more efficient and compact representation of complex structures while preserving visual quality comparable to NeRF. As a result, 3DGS has quickly gained traction as one of the leading solutions for view synthesis tasks.

However, real-world photos are often captured with low resolution (LR), making high-resolution Novel view synthesis (HRNVS) particularly challenging. LR images typically lack sufficient detail, limiting the ability to faithfully reconstruct high-frequency information and resulting in noticeable artifacts when synthesizing high-resolution (HR) target views~\cite{yu2024mip}. Therefore, 3DGS super-resolution (SR) methods become crucial for enabling reliable HRNVS.

Mainstream methods leverage image SR priors to guide 3DGS toward HRNVS. These approaches fall into two categories. The first category~\cite{feng2024srgs, yu2024gaussiansr,wan2025s2gaussian, xie2024supergs} relies heavily on single-image super-resolution (SISR) models. Although this improves the perceptual quality of individual images, these methods fail to fuse complementary information across views and lack cross-view consistency, resulting in poor recovery of high-frequency details in the synthesized views. The second category~\cite{shen2024supergaussian, ko2025sequence} leverages video SR techniques to overcome the shortcomings of the aforementioned methods. However, they rely on temporal continuity and alignment, whereas some multi-view datasets cannot be organized into coherent video streams, making direct adoption impractical.

Following prior studies while addressing their limitations, we focus on tackling 3DGS HRNVS problems by leveraging SR priors from multi-view observations. To this end, we propose a novel \textbf{M}ulti-\textbf{V}iew consistent 3D \textbf{G}aussian \textbf{S}platting \textbf{S}uper-\textbf{R}esolution (MVGSR) framework. Our approach aims at integrating multi-view information for 3DGS rendering with high-frequency details and enhanced consistency. However, using all views is impractical due to the high computational cost. Thus, we propose an Auxiliary View Selection method based on camera pose to select views most informative for target SR, making it applicable to arbitrarily organized datasets. During auxiliary view selection, we extract and reorganize intrinsic and extrinsic camera parameters for all input views. By computing spatial and directional camera similarity, we identify the most suitable auxiliary views for each target, enabling effective high-frequency detail enhancement.

Furthermore, we design a hierarchical multi-view SR network that utilizes auxiliary image information to enhance the view quality. The target and auxiliary features are first extracted through a Multi-View Feature Extraction Module with an epipolar-constrained attention mechanism to integrate auxiliary image features into the target image feature representation. By identifying geometry-consistent correspondence regions, this mechanism precisely combines relevant features in auxiliary views. The fused multi-scale features are progressively decoded and upsampled with SISR prior to reconstruct HR target images with improved geometric consistency and high-frequency details. The enhanced views together with the LR views are further used to optimize 3DGS, with an anti-aliasing subpixel loss.

Importantly, our method is the first to apply epipolar-constrained multi-view attention to 3DGS HRNVS, allowing efficient multi-view feature integration with lower computational cost, thus delivering substantially improved texture detail and structural consistency for 3DGS rendering. Experiments demonstrate that our method consistently achieves state-of-the-art (SOTA) 3DGS performance on both single-object and complex scene datasets.

In summary, our main contributions are as follows:
\begin{itemize}
\item We present MVGSR, a novel multi-view consistent SR framework that enables HR 3DGS scene reconstruction and enhances high-frequency details for NVS.
\item We propose an auxiliary view selection method based on camera poses from arbitrarily organized multi-view datasets to select views that are more informative for the target image SR.
\item We introduce a multi-view SR network equipped with an epipolar-constrained attention mechanism, which enables geometry-aware cross-view fusion, significantly improving spatial consistency and high-frequency detail in 3DGS rendering.
\item Experiments show that our approach provides SOTA performance in 3DGS rendering quality over prior methods, improving cross-view consistency and high-frequency details on both single-object and complex scene datasets.
\end{itemize}






\section{Related Work}
\subsection{Novel View Synthesis}
NVS aims to generate new images from arbitrary target viewpoints given several observed images from known perspectives, which is a key technology in areas such as 3D reconstruction and virtual reality. Classical methods like NeRF~\cite{niemeyer2022regnerf} implicitly represent the scene as a continuous volumetric field and achieve high-quality novel view generation through neural rendering. However, despite numerous subsequent efforts to enhance its performance~\cite{barron2021mip, hedman2021baking, lin2022efficient, liu2020neural}, NeRF-like methods remain constrained in practical applications due to their reliance on high-density sampling and the resulting inefficiency in real-time rendering. More recently, 3DGS~\cite{kerbl20233d} has emerged as a prominent explicit scene representation. Unlike NeRF, 3DGS describes the 3D structure with explicit Gaussian primitives and enables real-time, high-quality rendering through differentiable rasterization. Despite its advantages, early 3DGS methods suffer from sparse Gaussian distributions and insufficient texture details, resulting in a noticeable quality drop in HR rendering. Subsequent works~\cite{yu2024mip, fang2024mini, ye2024absgs, lu2024scaffold}, such as Mip-Splatting, have alleviated rendering artifacts via spatial smoothing in the 3D domain but still struggle to recover rich textures from low-resolution inputs.

\begin{figure*}[t]
\centering
\includegraphics[width=0.9\textwidth]{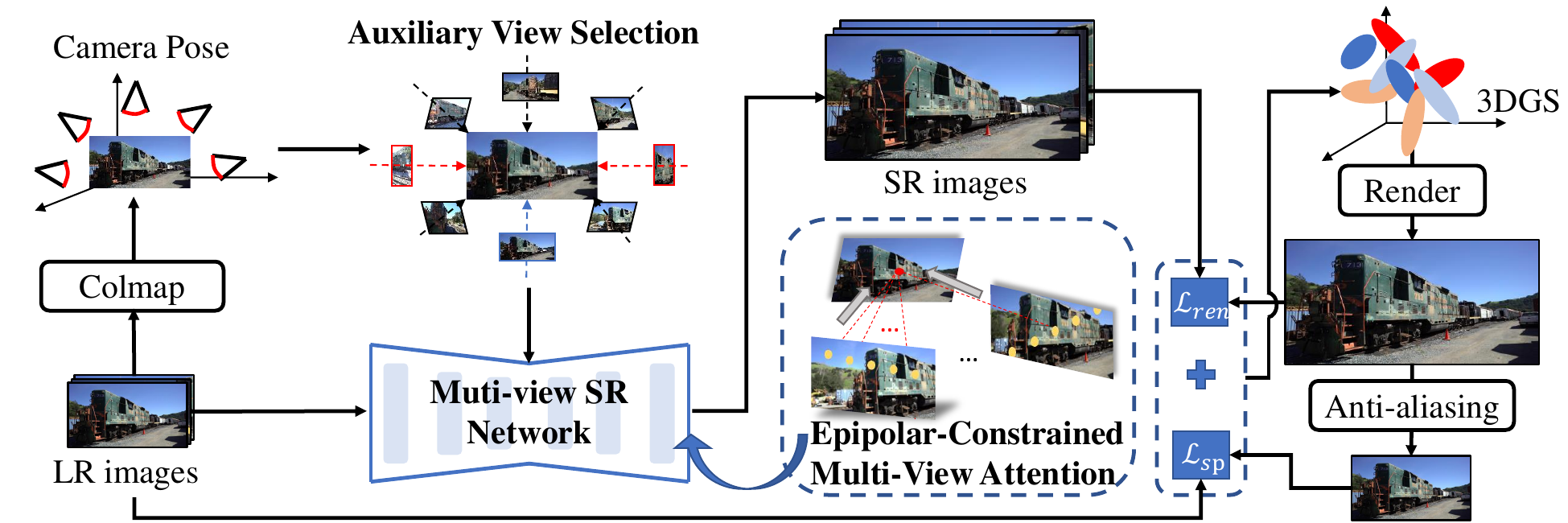} 
\caption{An overview of the proposed MVGSR pipeline. Given a set of LR images and their corresponding camera poses estimated via COLMAP, we first select auxiliary views based on camera pose. The selected auxiliary views, together with the target LR image, are fed into a multi-view SR network. It employs an epipolar-constrained multi-view attention mechanism to extract consistent and complementary high-frequency details from the auxiliary views. The resulting super-resolved images, together with the original LR images, are then used to jointly train the 3DGS.}
\label{pipeline}
\end{figure*}

\subsection{3D Scene Super-Resolution}
The objective of 3D scene SR is to reconstruct scenes capable of HR rendering from LR inputs. While SISR methods based on transformer architectures~\cite{wang2022uformer, liang2021swinir, zamir2022restormer} have achieved impressive results in 2D vision, they often produce texture inconsistencies across different views and are unable to exploit the complementary information available between multiple views. As a result, their effectiveness for 3D scene reconstruction is limited. Recent efforts have explored 3D SR approaches based on NeRF, such as NeRF-SR~\cite{wang2022nerf} and FastSR-NeRF~\cite{garbin2021fastnerf}. However, the performance of these methods is still restricted by the inherent modeling capabilities of NeRF and face challenges in recovering high-fidelity details.

With the rise of 3DGS, several studies have integrated SISR and 3DGS SR. SRGS~\cite{feng2024srgs}, for example, introduces external high-quality SISR models to provide texture priors, thereby enhancing the capability of Gaussian primitives to represent fine-grained details. GaussianSR~\cite{yu2024gaussiansr} leverages score distillation sampling to introduce large-scale generative image priors into 3D scene SR. SuperGS~\cite{xie2024supergs} adopts a two-stage coarse-to-fine framework, where an LR scene is first optimized and then refined using a pre-trained SISR model to enhance HR details. Although these methods attempt to improve cross-view consistency, their reliance on SISR makes it difficult to maintain texture consistency and recover high-frequency details across views. Another line of work, such as SuperGaussian~\cite{shen2024supergaussian} and SM~\cite{ko2025sequence}, applies video SR for 3DGS SR. However, SuperGaussian applies video SR to the rendered trajectories of pre-reconstructed 3DGS scenes, which tends to amplify reconstruction errors present in the LR input. SM relies on a pretrained video SR model and requires an additional view-ordering procedure to rearrange unordered multi-view inputs. However, such sorting is computationally intensive and resource-demanding, which limits its applicability in resource-constrained or large-scale multi-view scenarios.

To address these limitations, we propose MVGSR, a multi-view image-based 3DGS SR framework that avoids the texture inconsistencies and missing cross-view high-frequency details of SISR, while also eliminating error accumulation and poor generalization in video SR methods.
\section{Methods}

\subsection{Overview}
In this work, we propose a novel Multi-View consistent 3D Gaussian Splatting Super-Resolution (MVGSR) framework. The goal of our framework is to leverage multi-view images for enhancing high-frequency details and cross-view consistency, enabling high-quality scene reconstruction and NVS. As illustrated in \cref{pipeline}, we first use COLMAP~\cite{schonberger2016structure} to estimate the camera poses for a set of input views. For each input LR image $\mathrm{LR}_i$, our auxiliary view selection method identifies $n$ auxiliary views and their camera poses. These images and pose information are then fed into our designed multi-view SR network equipped with the epipolar-constrained multi-view attention to generate a super-resolved image $\mathrm{SR}_i$ that incorporates both multi-view consistency and high-frequency details. Using the super-resolved images together with the original LR observations as guidance, our framework trains 3DGS to exploit cross-view consistency and high-frequency complementary information from auxiliary views, resulting in substantially improved 3D reconstruction quality.

The following sections provide detailed descriptions of each module, including the auxiliary view selection method based on camera poses, the architecture of the multi-view SR network, the epipolar-constrained multi-view attention mechanism, and the loss functions used in our framework.

\begin{figure*}[t]
\centering
\includegraphics[width=0.95\textwidth]{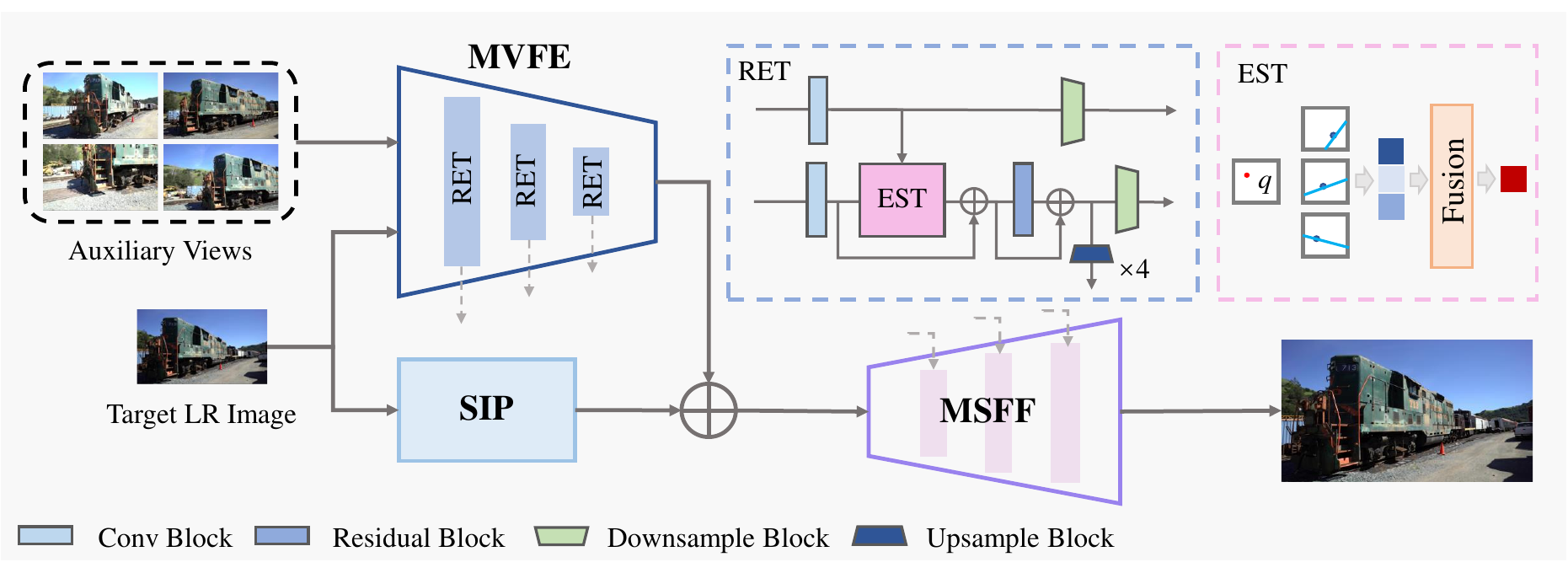} 
\caption{The architecture of the Multi-View SR Network. The whole network consists of the MVFE Module, the SIP Module, and the MSFF Module. The LR target and auxiliary images are taken as input for MVFE to extract multi-view features. The MVFE consists of 3 RET blocks at different scales, each integrated with an EST module employing epipolar-constrained multi-view attention. Combined with the single-image deep prior by the SIP module, the target image is effectively restored by fully fusing the single-image feature with the multi-view feature.}
\label{Network}
\end{figure*}

\subsection{Auxiliary View Selection Method Based on Camera Poses}
For integrating multi-view information for 3DGS rendering, we design an auxiliary view selection method based on camera pose information. Specifically, We first use COLMAP to obtain a set of accurate camera poses from the dataset and store them in a unified camera-to-world format, including all intrinsic and extrinsic parameters, which also facilitates the subsequent invocation of the multi-view SR network. Then, we use the camera poses to select views that provide rich complementary information for the target view. The core principle of the auxiliary view selection method is that, for any target view, its selected auxiliary views should satisfy the following three conditions:
1) The camera position should be closer to the scene center than the target camera so that the auxiliary view may provide finer details;  
2) There should be a certain degree of content overlap with the target view, which is the prerequisite for effective information supplementation;  
3) The auxiliary camera pose should not be too close to the target camera pose, otherwise the content will be redundant and less informative.

To this end, we compute both the camera position and the direction of each view. We first perform a filtering step based on two geometric constraints, and then rank the remaining candidates according to a mixed distance metric that combines spatial and directional similarity, finally selecting a fixed number of auxiliary views for further feature fusion. Specifically, assuming we have $n$ cameras (the $i$th camera position and direction are denoted as $P_i$ and $d_i$, respectively), the auxiliary view selection can be formulated as follows:

\textbf{Step 1: Candidate Filtering.}  
For each candidate view $j$, we retain only those that satisfy:

\begin{itemize}
    \item \textbf{Condition 1 (closer to scene center):}
    \begin{equation}
    d_i^T (P_j - P_i) > 0.
    \end{equation}
    That is, assuming the target camera $i$ faces the scene center, its direction forms an acute angle with the vector from $i$ to $j$.
    \item \textbf{Condition 2 (viewing cone overlap):}
    Let $\theta_{j, i\rightarrow{j}}$ denotes the angle between $P_j - P_i$ and $d_j$:
    \begin{equation}
    \sin \theta_{j, i\rightarrow{j}} \geq \frac{1}{2},
    \end{equation}
    indicating sufficient overlap between the view $i$ and $j$.
\end{itemize}

\textbf{Step 2: Distance Computation and Ranking.}  
For the filtered candidates from Step 1, we calculate the position distance $D_{pos}(ij)$ and the camera direction distance $D_{dir}(ij)$ to the target camera $i$:

\begin{align}
D_{pos}(ij) &= \| P_i - P_j \| ,
\end{align}
\begin{align}
D_{dir}(ij) &= 1 - \frac{d_i^T d_j}{\|d_i\| \cdot \|d_j\|} .
\end{align}

According to the above, the final distance between the target camera $i$ and the candidate camera $j$ is computed as
\begin{equation}
D_{ij} = 
\begin{cases}
\lambda_{pos} D_{pos}(ij) + (1-\lambda_{pos}) D_{dir}(ij), \\
\quad \text{if } d_i^{\top} (P_j - P_i) > 0 \text{ and } \sin \theta_{j, i\rightarrow{j}} \geq \frac{1}{2} \\
\infty, \qquad \text{otherwise} ,
\end{cases}
\end{equation}
where $\lambda_{pos}$ is a balancing weight between two distances; empirically, we set $\lambda_{pos}=0.5$.

\textbf{Step 3: Final Selection.}  
Finally, we sort the candidate views according to $D_{ij}, j=1,2,\ldots, n$ and select $N_{ref}$ auxiliary views by sampling one view every $l$ positions along the sorted list, rather than simply choosing the top-$N_{ref}$ closest views. This step ensures the third condition that the selected auxiliary views are both informativeness and diverse. For further explanation and details on our auxiliary view selection, please refer to the Appendix.

\subsection{Multi-View Super-Resolution Network Architecture}

As illustrated in \cref{Network}, our Multi-View SR Network consists of three modules: the Multi-View Feature Extraction Module, the Single-Image Prior Module, and the Multi-Scale Feature Fusion Module. These three components cooperate to achieve high-quality reconstruction of the LR target image from auxiliary LR views.

\textbf{Multi-View Feature Extraction (MVFE) Module.}
This module is responsible for extracting informative features from multiple auxiliary images and integrating them with the target image’s feature. The feature extractor consists of three Residual Epipolar Transformer (RET) blocks, operating at different spatial scales. At each block, shallow convolutional layers $f$ extract LR features from both the target and auxiliary images. These features are processed by an Epipolar-Guided Spatial Transformer (EST), which identifies the target–auxiliary correlation by explicitly exploiting their geometry-aware projection consistency (see Section 3.4 for more details). The propagation process at each stage is formulated as
\begin{align}
x_j =\ & f^j(x_j) + \mathrm{EST}^j\Big(
         f^j(x_j),\ f^j(r_1(x_j)), f^j(r_2(x_j)),\notag\\&
       \ldots,\ 
         f^j(r_{N_{ref}}(x_j))
         \Big), \quad j=1,2,3,
\end{align}
where $x_j$ is the input of Layer $j$, and $f^j(\cdot)$ denotes the $j$-th convolutional layer, and $r_i(x_j)$ represents  the $i$-th auxiliary view of the $x_j$. The extracted features are further refined by a residual block: $x_{j+1} = \mathrm{Res^j}(x_j)$.

\textbf{Single-Image Prior (SIP) Module.}
Considering that large-scale pre-trained SISR models inherently possess rich priors for single-image detail recovery, we employ a feature extraction module based on SISR prior to obtain deep features from the LR target image. Specifically, we adopt the SwinIR feature extraction module based on Swin-Attention and initialize it with the corresponding pre-trained weights \cite{liang2021swinir}. This module provides a robust foundational representation for subsequent multi-scale feature fusion while reducing training costs. 

\textbf{Multi-Scale Feature Fusion (MSFF) Module.}
At each scale, the auxiliary feature from the corresponding level of the MVFE module is concatenated with the target feature along the channel dimension and fused via convolutions. The fused features are then upsampled to the next scale. This process is repeated across multiple levels to progressively reconstruct HR features, enabling high-quality restoration from the target image feature and multi-view feature. See the Appendix for more model details.

\begin{figure}[t]
\centering
\includegraphics[width=0.9\columnwidth]{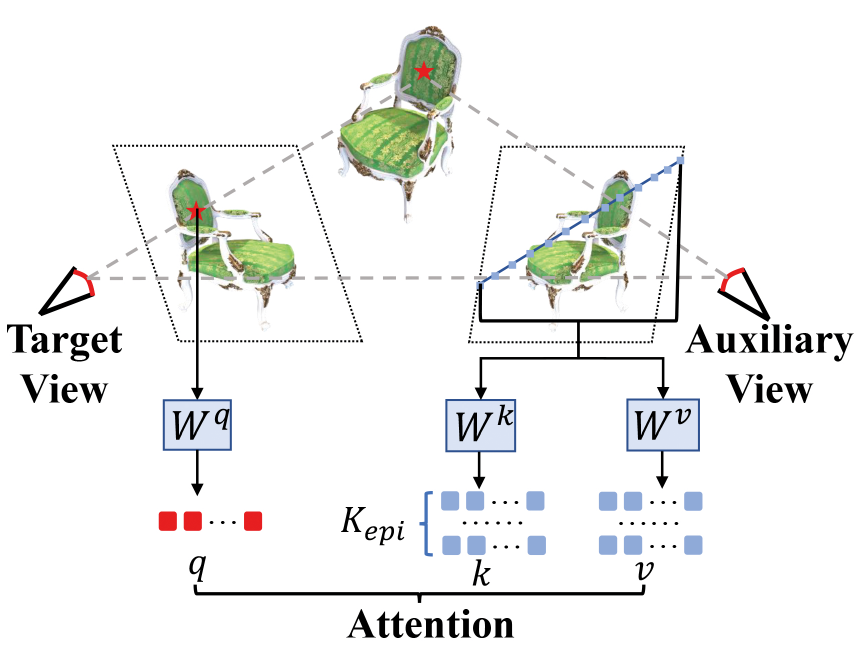} 
\caption{Epipolar-Constrained Multi-View Attention}
\label{epi}
\end{figure}

\subsection{Epipolar-Constrained Multi-View Attention}
\label{sec:epipolar_attention}
The introduced epipolar-constrained multi-view attention mechanism is the core of EST. By identifying geometry-consistent correspondence regions, it precisely locates relevant auxiliary features, enabling the system to aggregate richer information from a larger number of auxiliary views within limited computational budgets. As shown in \cref{epi}, the core idea is to leverage multi-view geometry by projecting each query point in the target view onto its corresponding epipolar line in the auxiliary views \cite{hartley2003multiple}, where the point corresponding to the query must lie. In this way, we restrict attention computation to geometrically valid regions and reduce irrelevant information exchange.

\textbf{Epipolar Region Computation.} In the feature level, given a pixel $x_i$ in the target view and the intrinsic and extrinsic parameters of both target and auxiliary cameras, the corresponding epipolar line $l_j$ in the auxiliary view is computed as
\begin{equation}
l_j = \mathbf{F}_{ij} \, \tilde{x}_i,
\end{equation}
where $\tilde{x}_i$ denotes the homogeneous coordinates of $x_i$, and $\mathbf{F}_{ij}$ is the fundamental matrix between the target and reference views~\cite{hartley2003multiple}.



\textbf{Epipolar Line Sampling and Attention Computation.}  
For $x_i$, we uniformly sample $K_{epi}$ candidate points $\{y_j^k\}_{k=1}^K$ along each epipolar line $l_j$ in the reference view. The features at these locations are extracted as key and value matrices $K_j$, $V_j$ for the attention computation. Attention is computed over these sampled points:
\begin{equation}
\alpha_{i, j}=\operatorname{Softmax}\left(\frac{q_i \cdot K_j}{\sqrt{d}}\right),
\end{equation}
where $q_i$ is the query feature from the target view.

The aggregated cross-view feature is
\begin{equation}
f_{i,j}^{\text{epi}} = \alpha_{i, j} V_{j}.
\end{equation}

Finally, all aggregated features from auxiliary views are integrated using a self-attention mechanism, resulting in the final enriched representation of $x_i$.

\subsection{Loss Functions}
For training the SR network, we adopt a composite loss that includes reconstruction loss $\mathcal{L}_{\mathrm{rec}}$, and perceptual loss $\mathcal{L}_{\mathrm{per}}$:
\begin{equation}
\mathcal{L}_{\mathrm{SR}} = \mathcal{L}_{\mathrm{rec}} + \lambda_{\mathrm{per}}\mathcal{L}_{\mathrm{per}}.
\end{equation}
Specifically, we use the $\ell_1$-norm for $\mathcal{L}_{\mathrm{rec}}$. The perceptual loss $\mathcal{L}_{\text{per}}$~\cite{johnson2016perceptual} is computed using the VGG19~\cite{simonyan2014very}.

For 3DGS training, traditional methods optimize only HR reconstruction against the ground truth, which in SR settings neglects interactions with the LR inputs and can lead to degradation of scene structure. To mitigate this, we introduce a sub-pixel loss into the 3DGS training process, following prior work~\cite{feng2024srgs}. Unlike previous methods, we apply an anti-aliased bicubic downsampling scheme to the rendered images, which more faithfully preserves high-frequency details and thus provides more reliable supervision. The sub-pixel loss $\mathcal{L}_{\mathrm{sp}}$ is computed between these downsampled renders and the LR ground-truth images. The overall objective of 3DGS is
\begin{equation}
\mathcal{L}_{\mathrm{3DGS}}
=\lambda_{\mathrm{ren}}\mathcal{L}_{\mathrm{ren}}+(1-\lambda_{\mathrm{ren}})\mathcal{L}_{\mathrm{sp}},
\end{equation}
where $\mathcal{L}_{\mathrm{ren}}$ denotes the SR 3DGS rendering loss.
\section{Experiment}

\begin{table}[t]
\caption{Quantitative comparison on NeRF Synthetic ×4 and ×2 (8 views). The numbers marked with \dag \space  are sourced from their respective paper. The best and second best entries are marked in red and orange, respectively.}
\label{synthetic}
\centering
\begin{tabular}{lccc}
\toprule
\textbf{Methods} & \textbf{PSNR↑} & \textbf{SSIM↑} & \textbf{LPIPS↓}\\
\midrule
\multicolumn{4}{c}{\textbf{400×400 → 800×800}} \\
\hline
Bicubic-3DGS\shortcite{kerbl20233d}   & 30.97 & 0.9540 & 0.0581   \\
SwinIR-3DGS\shortcite{liang2021swinir}    & 32.57 & 0.9640 & 0.0374     \\
Mip-Splatting\shortcite{yu2024mip}  &  31.37   &   0.9572   &   0.0485   \\
\hline
NeRF-SR\shortcite{wang2022nerf} & 30.08 & 0.9391 & 0.0501\\
\hline
SRGS\shortcite{feng2024srgs}         & \second{32.67} & \second{0.9643} & \second{0.0371}  \\
Ours           & \best{33.01}& \best{0.9655} & \best{0.0368}   \\
\hline
\multicolumn{4}{c}{\textbf{200×200 → 800×800}} \\
\hline
Bicubic-3DGS\shortcite{kerbl20233d} & 27.54 & 0.9149 & 0.1158 \\
SwinIR-3DGS\shortcite{liang2021swinir} & 30.76 & \second{0.9498} & 0.0561 \\
Mip-splatting\shortcite{yu2024mip} & 27.54 & 0.9146 & 0.1162 \\
\hline
NeRF-SR\shortcite{wang2022nerf} & 28.45 & 0.9211 & 0.0758 \\
FastSR-NeRF\dag\shortcite{lin2024fastsr} & 30.47 & 0.9440 & 0.0750 \\
\hline
SRGS\shortcite{feng2024srgs} & 30.83 & 0.9480 & \second{0.0560}\\
SuperGS\dag\shortcite{xie2024supergs} & \second{30.89}	& 0.9490 & 0.0560 \\
GaussianSR\dag\shortcite{yu2024gaussiansr} & 28.37 & 0.9240  & 0.0870\\
SuperGaussian\dag\shortcite{shen2024supergaussian} & 28.44 & 0.9230 & 0.0670\\
SM\shortcite{ko2025sequence} & 28.05 & 0.9377 & 0.0627\\
Ours & \best{31.11}& \best{0.9503} & \best{0.0550}\\
\hline
HR-3DGS         & 33.32 & 0.9749 & 0.0239 \\
\bottomrule
\end{tabular}
\end{table}

\subsection{Setup}
All experiments are conducted on a single NVIDIA RTX 4090 GPU. During the training phase of our SR network, we set the total number of iterations to 200,000, a batch size of 2, and the number of epipolar sampling points $K_{epi}$ to 64, 32, and 16 for each block. The number of auxiliary images is fixed at 4, with selecting step size 2. In addition, we decay the learning rate from 1e-4 to 1e-7 in a cosine annealing way. The 3DGS training phase follows the original settings; to ensure fairness, all methods are evaluated under identical data and 3DGS training configurations unless otherwise stated. For more hyperparameter analysis and selection criteria, please see the Appendix. To comprehensively evaluate the robustness of our approach, we conduct experiments on the following datasets:

\begin{figure*}[t]
\centering
\includegraphics[width=1\textwidth]{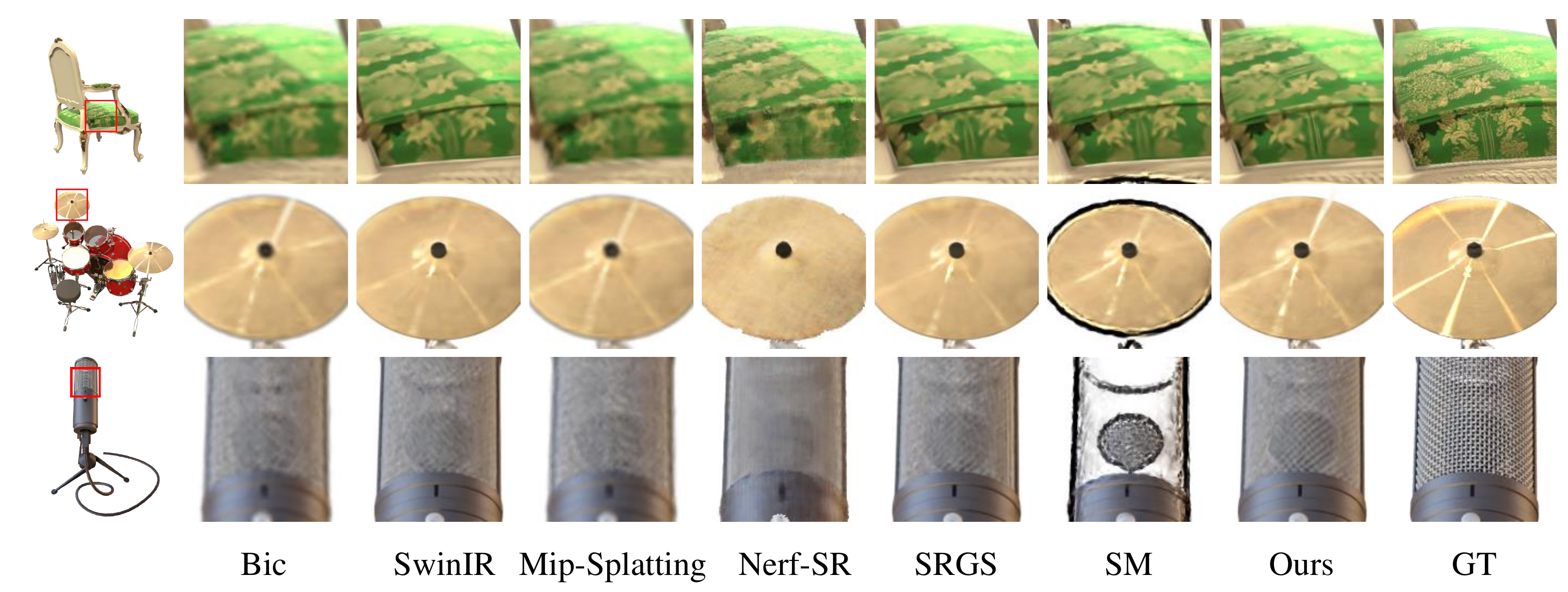} 
\caption{Qualitative comparisons on NeRF Synthetic ×4 datasets. MVGSR produces more visually appealing results, successfully capturing high-frequency details and textures. Best viewed at screen!}
\label{fig-syn}
\end{figure*}

\begin{figure}[t]
\centering
\includegraphics[width=1\columnwidth]{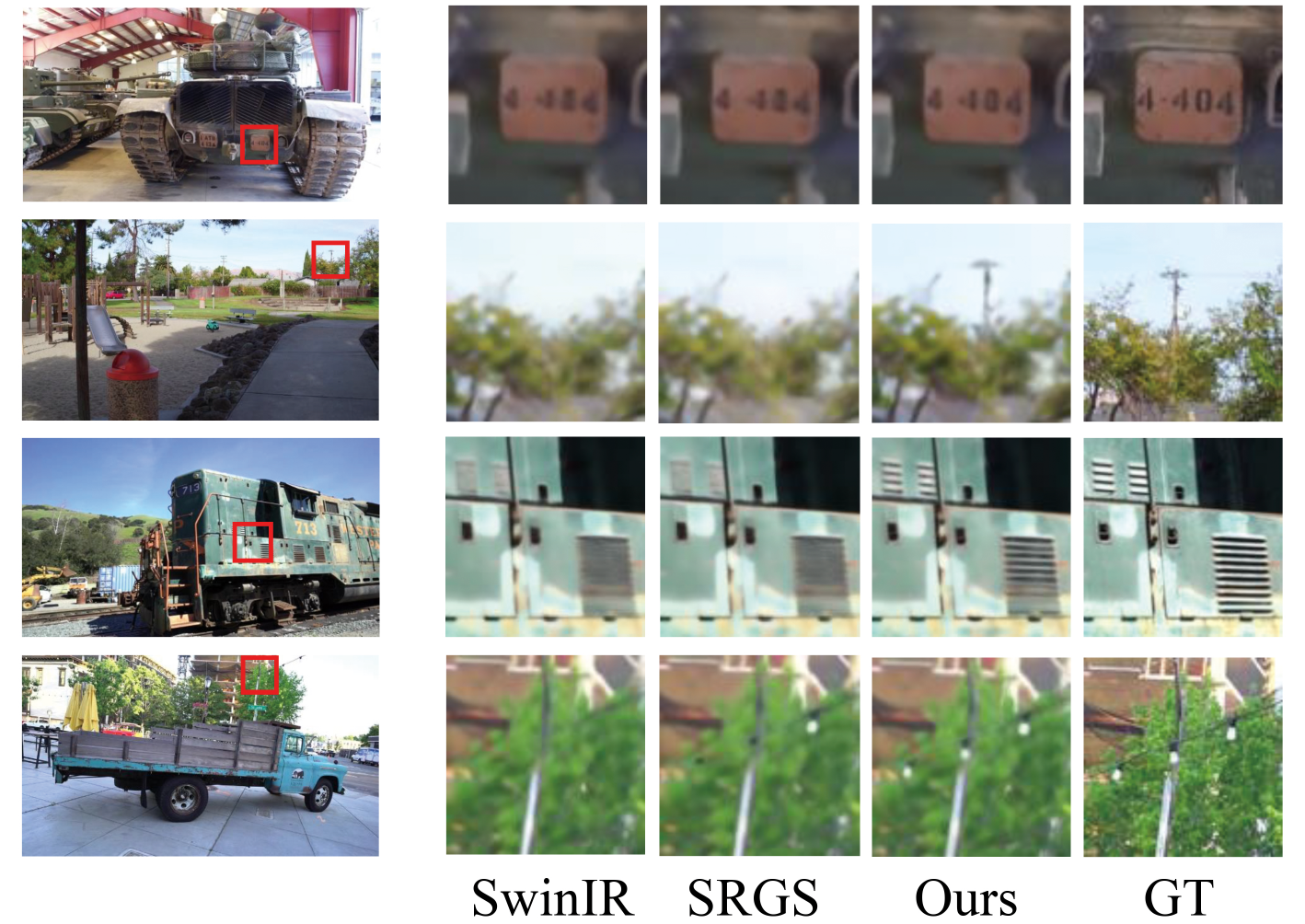} 
\caption{Qualitative comparisons on Tanks \& Temples dataset of 240×135 → 960×540 task. MVGSR consistently restores coherent structures and intricate details. Best viewed at screen!}
\label{fig-tanks}
\end{figure}

\textbf{Tanks \& Temples Dataset}~\cite{knapitsch2017tanks} is a real-world dataset. We select four scenes for testing and use the remaining scenes for training our SR network. The original resolution of the images is 1920×1080. For efficient training, we resize the images to 960×540 and test the performance at downscaled x2 and x4 resolutions. On the four testing scenes, we select every 8 images for 3DGS testing.

\textbf{Mip-NeRF 360 Dataset}~\cite{barron2022mip} comprises 9 real-world scenes, with 5 outdoors and 4 indoors, each containing a complex central object or area with a detailed background. We downsample the training views by a factor of $\times 4$ as low-resolution inputs and directly apply our multi-view SR network for testing. Following prior work, we select every 8 images for 3DGS testing.

\textbf{NeRF Synthetic Dataset}~\cite{wang2022nerf} is a collection of 8 single-object scenes, each with images at a resolution of 800×800. We use 100 images for 3DGS training and 200 test images for evaluation. All images are downsampled by a factor of 4 to generate the LR inputs.

For quantitative evaluation, we adopt the following metrics: Peak Signal-to-Noise Ratio (PSNR)~\cite{huynh2008scope}, Structural Similarity Index (SSIM)~\cite{wang2004image}, and Learned Perceptual Image Patch Similarity (LPIPS)~\cite{zhang2018unreasonable}. Since our objective is to improve 3DGS reconstruction quality rather than standalone image SR, we treat PSNR—which measures deviations from the original scene—as the primary evaluation metric. All results are obtained through rendering after 3DGS reconstruction, and the metrics are computed by comparing the rendered images with the ground-truth (GT) images.

\begin{table}[t]
\caption{Quantitative comparison on Tanks \& Temples Dataset (4 views) at two resolution scales. The numbers marked with $*$ contain only two scenarios: Truck and Train. The numbers marked with \dag \space  are sourced from their respective paper. The best and second best entries are marked in red and orange, respectively.}
\centering
\begin{tabular}{lccc}
\toprule
\textbf{Method} & \textbf{PSNR↑} & \textbf{SSIM↑} & \textbf{LPIPS↓} \\
\midrule
\multicolumn{4}{c}{\textbf{240×135 → 960×540}} \\
\hline
Bicubic-3DGS\shortcite{kerbl20233d}   & 24.45 & 0.7699 & 0.3580 \\
Mip-Splatting\shortcite{yu2024mip}  & 24.42 & 0.7743 & 0.3573 \\
SwinIR-3DGS\shortcite{liang2021swinir}    & \second{25.57} & \second{0.8385} & \second{0.2772} \\
SRGS \shortcite{feng2024srgs}          & 25.38 & 0.8281 & 0.2874 \\
Ours           & \best{25.75} & \best{0.8406} & \best{0.2705} \\
\hline
HR-3DGS        & 26.63 & 0.8921 & 0.1888 \\
\hline
\multicolumn{4}{c}{\textbf{480×270 → 1920×1080}} \\
\hline
Bicubic-3DGS\shortcite{kerbl20233d}   & 24.23 & 0.7681 & 0.3664   \\
Mip-Splatting\shortcite{yu2024mip}  &  24.22   &   0.7738   &   0.3724   \\
SwinIR-3DGS\shortcite{liang2021swinir}    & 24.77 & \second{0.8102} & \second{0.3217}     \\
SRGS\shortcite{feng2024srgs}         & \second{24.79} & 0.8052 & 0.3315  \\
Ours           & \best{24.90}& \best{0.8110} & \best{0.3205}   \\
\hline
SuperGS\dag$*$\shortcite{xie2024supergs}     & 21.19 & 0.6950 & 0.3640 \\
Ours$*$        & \textbf{23.31} & \textbf{0.8313} & \textbf{0.3258}   \\
\hline
HR-3DGS         & 25.30 & 0.8545 & 0.2781 \\
\bottomrule
\end{tabular}
\label{tanks}
\end{table}

\subsection{Quantitative and Qualitative Comparisons}
To rigorously validate the effectiveness of our proposed MVGSR, we conduct extensive comparisons against a range of existing approaches, including NeRF-based methods (NeRF-SR~\cite{wang2022nerf} and FastSR-NeRF~\cite{lin2024fastsr}) and 3DGS-based methods (SRGS~\cite{feng2024srgs}, SuperGS~\cite{xie2024supergs}, GaussianSR~\cite{yu2024gaussiansr}, SuperGaussian~\cite{shen2024supergaussian}, and SM~\cite{ko2025sequence}). Due to the unavailability of source code for some approaches, we directly report results from their original papers under identical configurations for fairness. Additionally, we include comparisons with three baselines: Bicubic-3DGS, SwinIR-3DGS, and Bicubic-Mip-Splatting~\cite{yu2024mip}. For the complete results on the Mip-NeRF 360 Dataset, as well as the additional results on the NeRF Synthetic Dataset and the Tanks \& Temples Dataset, please refer to the Appendix.


\textbf{Quantitative Results.}
As shown in \cref{synthetic}, our method achieves the best overall balance between quantitative accuracy and perceptual quality, outperforming existing 3DGS-based SR approaches in both ×4 and ×2 settings. The improvements in PSNR and LPIPS demonstrate its stronger ability to reconstruct high-frequency details and maintain cross-view consistency. The outstanding results on the object-centric NeRF Synthetic dataset confirm that our method can achieve high-fidelity reconstruction and consistent SR performance at the single-object level.

As illustrated in \cref{tanks}, MVGSR consistently surpasses all existing methods across evaluation metrics, indicating its robustness in integrating multi-view information to recover fine-grained textures and high-frequency structures even in complex large-scale scenes.


\textbf{Qualitative Results.}
Visual comparisons presented in \cref{fig-syn} and \cref{fig-tanks}, highlight MVGSR’s superior capability in recovering high-frequency details and fine textures. Conventional 3DGS-based methods commonly exhibit significant artifacts. Despite SwinIR enhancing local details, independently upscaling each view introduces cross-view inconsistencies, which ultimately cause noticeable detail loss in the rendered 3DGS views. Meanwhile, Mip-Splatting struggles to recover details due to insufficient high-frequency information, and NeRF-SR produces blurry reconstructions lacking precise textures. The SRGS method also suffers from blur and detail loss. Additionally, SM demonstrates reconstruction failures under white-background conditions, manifesting as substantial black artifacts. Conversely, MVGSR consistently restores coherent structures and intricate details while significantly mitigating visual artifacts prevalent in competing methods.

Across all tested configurations, MVGSR consistently surpasses existing approaches, delivering superior perceptual quality and quantitative results. By aggregating rich information from auxiliary views, it effectively enhances rendering cross-view consistency and reconstructs fine-grained and realistic details. The results further demonstrate strong generalization across diverse datasets.
\begin{table}[t]
\caption{Ablation studies on Tanks \& Temples Dataset (4 views, 240×135 → 960×540). ``Auxiliary → Near": use nearest-neighbor views instead of the auxiliary selection; ``Auxiliary → Random": use random views instead; ``Epi → Cross": replace epipolar attention with cross-attention.}
\centering
\begin{tabular}{llll}
\toprule
\textbf{Methods} & \textbf{PSNR↑} & \textbf{SSIM↑} & \textbf{LPIPS↓}\\
\midrule
MVGSR & \textbf{25.72} & \textbf{0.8404} & \textbf{0.2710} \\
Auxiliary → Near & 25.58 & 0.8384 & 0.2734\\
Auxiliary → Random  & 25.61 & 0.8384 & 0.2728\\
Epi → Cross & 25.56 & 0.8387 & 0.2738\\
\bottomrule
\end{tabular}
\label{ablation}
\end{table}

\begin{figure}[t]
\centering
\includegraphics[width=0.7\columnwidth]{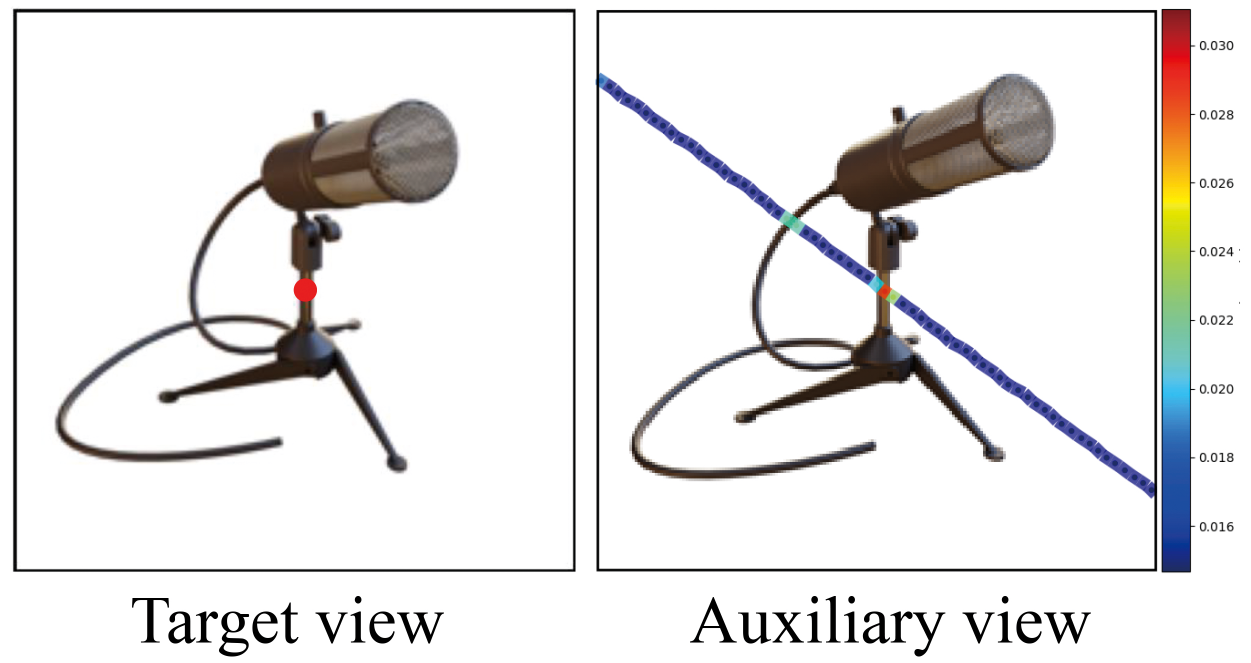} 
\caption{Attention distribution on the epipolar line. The query point (red dot) in the target view (left) attends to its corresponding regions along epipolar line in the auxiliary view (right), visualized as a heatmap.}
\label{epi_ver}
\end{figure}

\subsection{Ablation Studies and Analysis}
To assess the effectiveness of MVGSR components, comprehensive ablation studies are conducted. To facilitate efficient and fair comparison, each variant is trained for 70000 iterations with consistent training settings, as summarized in \cref{ablation}. The results verify that both the epipolar attention module and auxiliary view-selection strategy contribute to performance enhancements. Furthermore, replacing conventional cross-attention with epipolar attention lowers the computational cost from $O(N^2)$ to $O(N)$. This not only reduces memory consumption, but also enhances performance, since epipolar attention accurately identifies geometry-consistent feature regions for correspondence.

\begin{figure}[t]
\centering
\includegraphics[width=0.9\columnwidth]{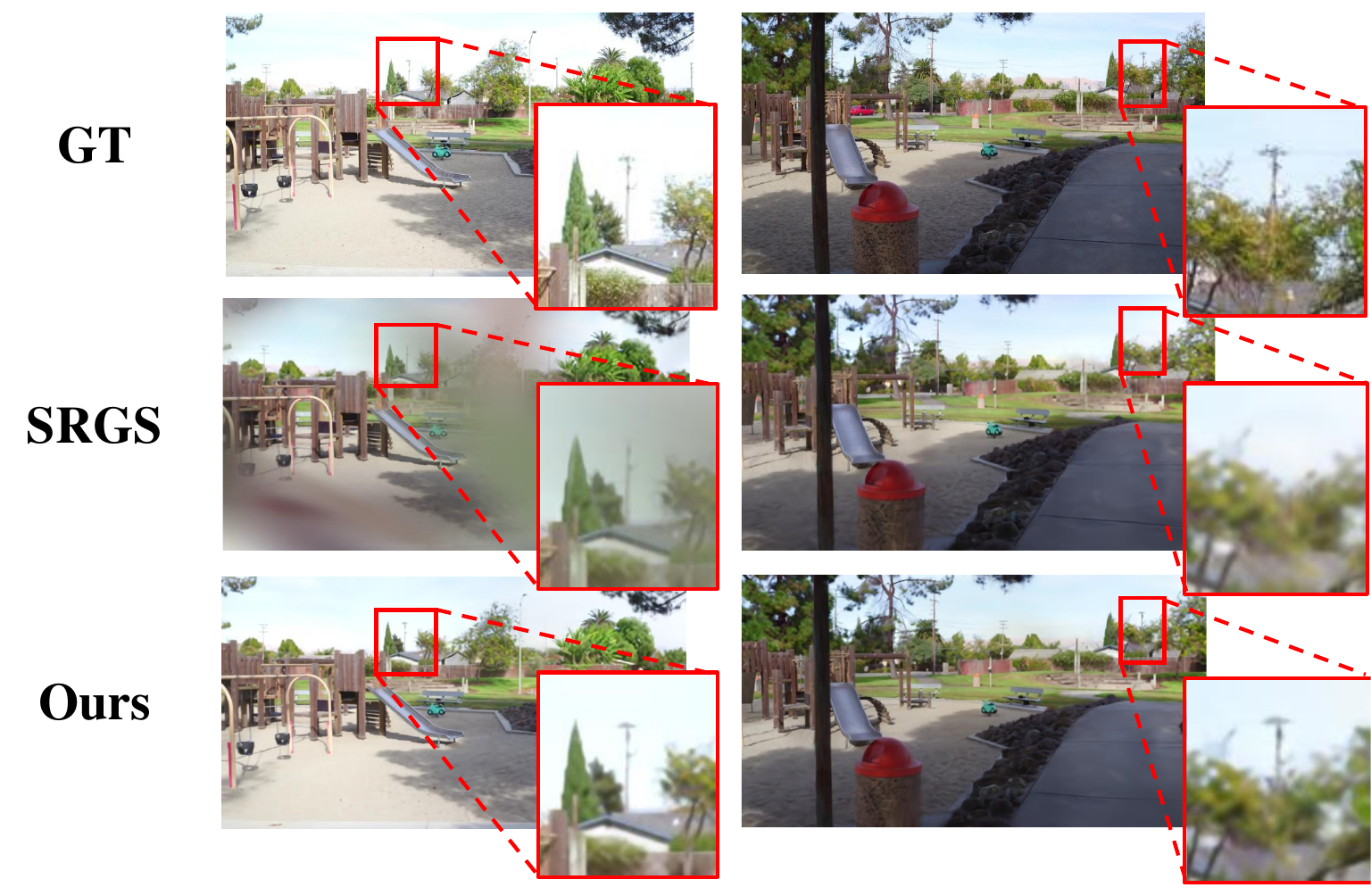} 
\caption{Comparison of cross-view consistency among different methods, along with failure cases caused by inconsistency.}
\label{fig_consist}
\end{figure}

To verify whether the epipolar-constrained multi-view attention can effectively identify corresponding information from auxiliary views, we visualize the attention distribution of the trained Network along the computed epipolar line. As shown in \cref{epi_ver}, our approach successfully attends to the corresponding regions of the query points, assigning them the highest attention scores. This demonstrates the capability of our approach to effectively leverage cross-view information for enhanced consistency and complementary image details. We also observe notable attention responses in regions that share similar texture patterns with the query point, suggesting that our epipolar attention also captures non-local similarities across views, further enriching the visual content. The cross view consistency is further demonstrated in \cref{fig_consist}. SRGS causes cross-view inconsistencies (e.g., objects appearing in one view but missing in another) because it independently super-resolves each view without geometric guidance. In contrast, our method aggregates geometrically aligned cues from auxiliary views, maintaining coherent structures across viewpoints.


We also compare the memory consumption of our auxiliary-view selection with the view-reordering approach employed in SM~\cite{ko2025sequence} in the bicycle scene of Mip-NeRF 360 Dataset. SM requires 25.94 GB of memory to process long video sequences for reordering. In contrast, our method uses only 0.46 GB, as the auxiliary-view selection relies only on camera poses rather than the images. This design enhances the generality and applicability of our method in memory-limited environments where video-based methods are infeasible, and making it suitable for large-scale, large-scene image datasets. More comparisons regarding method performance are provided in the Appendix.

\section{Conclusion}

In this paper, we introduce MVGSR, which integrates multi-view information to enhance 3DGS rendering with high-frequency details and improved consistency. We first propose an Auxiliary View Selection method based on camera poses, enabling adaptation to arbitrarily organized datasets without requiring view reordering. We also introduce an epipolar-constrained multi-view attention mechanism for 3DGS SR, which serves as the core of our network and aggregates geometrically consistent information from multiple auxiliary views. This design strengthens both geometric consistency and detail fidelity in 3DGS reconstruction. Experiments show that MVGSR achieves SOTA performance across 3DGS SR benchmarks. We believe MVGSR provides valuable insights for future research.

{
    \small
    \bibliographystyle{unsrt}
    \bibliography{main}
}


\end{document}